\documentclass[fleqn,10pt]{wlscirep}
\usepackage[utf8]{inputenc}
\usepackage[T1]{fontenc}
\usepackage{xcolor}
\usepackage{subcaption}

\title{\textbf{Massively Parallel Reinforcement Learning with a Chaotic Reconfigurable Clockless Chip}}

\author[1,*]{Eric Oliveira Gomes}
\author[1,*]{Damien Rontani}
\affil[1]{CentraleSup\'elec and Universit\'e de Lorraine, LMOPS UR4423 Laboratory, Metz F-57070, France}
\affil[*]{eric.oliveira-gomes@centralesupelec.fr - damien.rontani@centralesupelec.fr}


\begin{abstract}
Hardware accelerators based on physical dynamical systems offer an attractive route toward energy-efficient reinforcement learning applications. However, their scalability is challenging because it requires many statistically independent entropy sources. Here, we introduce a quasi-analog decision-making architecture based on asynchronous Boolean networks (or lattices) implemented on a clockless reconfigurable chip. Each node in the network consists of a single logic element that acts as an autonomous entropy source. This architecture gives rise to distributed Boolean chaos, in which a spatially coupled network generates parallel streams of chaotic Boolean transitions with very low statistical dependence between nodes. We experimentally demonstrate parallel decision-making on a 1024-armed bandit problem, which is beyond the scale of previous hardware implementations, while significantly improving power-law scaling performance. Separately, we scale the proposed entropy source to 5120 parallel channels, yielding an aggregate sample generation rate of 2.14~TS/s. Our solution is implemented on a commercial reconfigurable CMOS chip and offers high integration density and ease of programmability. Our results pave the way for using distributed Boolean chaos as a valuable hardware substrate for large-scale reinforcement learning and for the development of fully integrated, high-throughput decision-making accelerators.
\end{abstract}

\begin{document}
\flushbottom
\maketitle

\thispagestyle{empty}

\section*{Introduction}
Unconventional physics-based computing has emerged as an alternative to digital architectures to overcome their limitations and achieve high-bandwidth and energy-efficient information processing~\cite{jaeger_towards_2021, finocchio_roadmap_2024}. Limitations in digital design stems in part from the fundamental limits of circuit density in semiconductor technologies, which are driving the end of Moore's law~\cite{khan_science_2018}. As physically scaling digital circuits becomes increasingly harder, novel approaches exploiting a various physical phenomena have been proposed. One of these approaches is the use of deterministic chaos as a resource for computation, which has been shown to be applicable to reinforcement learning problems.

Reinforcement learning (RL) is a subfield of artificial intelligence in which an agent is trained to maximize an accumulated reward value obtained throughout a sequence of actions. The learning procedure happens through trial and error, with the agent gradually adapting its policy according to the rewards received from its interactions with the environment~\cite{sutton2018reinforcement}. Within this context, we focus on the multi-armed bandit (MAB) problem, a fundamental formulation that isolates the exploitation-exploration trade-off central to RL.

The MAB problem is defined by an environment composed of slot machines, also called arms, which have fixed win probabilities unknown to the agent. The agent interacts with the environment by choosing a slot machine, which will either yield a win or a loss through a Bernoulli trial defined by its win probability; each win provides a reward, while losses provide no reward. Given this setup, the agent must employ an algorithm allowing it to maximize the accumulated reward in a finite number of plays. This establishes a trade-off between exploration and exploitation, since the agent must be capable of exploration by trying less-known slot machines to gather information about their expected rewards, but should also be capable of exploitation by selecting among the best-known machines in order to maximize its reward. Algorithms developed for the MAB problem address this trade-off through different mathematical approaches. A few classic approaches include greedy methods, e.g. $\epsilon$-greedy~\cite{sutton2018reinforcement, auer_finite-time_2002}; upper confidence bound methods, e.g. UCB1 and UCB1-tuned~\cite{auer_finite-time_2002}; and Bayesian methods, e.g. Thompson sampling~\cite{thompson_likelihood_1933, chapelle_empirical_2011}. Beyond these algorithms, physics-based methods exploiting chaotic dynamics have also been proposed.

Chaos is a phenomenon in dynamical systems characterized by a sensitive dependence on initial conditions~\cite{strogatz2024nonlinear}. This property has been found to be exploitable in a variety of tasks, including secure communications~\cite{argyris_chaos-based_2005, garcia-ojalvo_spatiotemporal_2001, vanwiggeren_communication_1998}, simulation with Monte Carlo methods~\cite{yang_montecarlo_2013, umeno_chaotic_2000}, and random number generation~\cite{uchida_fast_2008, rosin_ultrafast_2013, kim_chiprng_2021, shen_harnessing_2023}.
In particular, for the MAB problem it has been applied to address the sampling bottleneck, through the use of chaotic signals as entropy sources in place of pseudo-random number generators. This approach has been shown to yield faster convergence than sampling from random distributions when applied to the tug-of-war (TOW) algorithm~\cite{naruse_ultrafast_2017}. Here, we propose Boolean chaos as an effective parallelizable entropy source for the TOW algorithm.

The TOW algorithm was initially proposed as a model inspired by the behavior of the \textit{P. polycephalum} amoeba, which collects environmental information by simultaneously extending multiple branches of its body, while maintaining a constant intracelular-resource volume~\cite{kim_tug-of-war_2010}. The nonlocal coupling between branches established by this volume conservation constraint allows for a natural handling of the exploration-exploitation trade-off. In practice, this volume conservation rule can be generalized to various conserved quantities in physical systems, which in theory enables its implementation on a wide variety of substrates~\cite{kim_efficient_2015}. A great number of TOW approaches have been developed based on photonics devices, exploiting high bandwidth chaotic signals achievable with optical systems. Photonic implementations have been demonstrated with single photons~\cite{naruse_single-photon_2015, naruse_single_2016}, quantum dots~\cite{naruse_decision_2014}, chaotic laser output intensities~\cite{naruse_ultrafast_2017, naruse_scalable_2018, morijiri_decision_2022}, multimode laser dynamics~\cite{homma_chip_2019, iwami_controlling_2022, iwami_experimental_2024, shao_harnessing_2025}, coupled laser networks~\cite{mihana_decision_2019, mihana_laser_2020, han_generation_2020, kotoku_asymmetric_2024}, optical spatiotemporal chaos~\cite{morijiri_parallel_2023}, and chaotic microcombs~\cite{shen_harnessing_2023}. These methods show promising results for solving the MAB problem, achieving shorter time-to-convergence than conventional software solutions. However, they rely on large experimental setups, requiring substantial hardware to scale the number of parallel arms, as well as complementary electronics to convert optical signals to the electronic domain for further processing. In contrast, we demonstrate that Boolean chaos enables on-chip, high-bandwidth chaos generation, with each arm's entropy source realized with a single logic gate. This minimal footprint allows for scaling to a massively parallel number of arms, while remaining natively compatible with standard CMOS electronics.

In this work, we study chaotic dynamics emerging from autonomous Boolean networks (ABNs). ABNs are networks of digital logic components that operate without a master clock, resulting in dynamics that depend on the physical properties of their components and unfold in continuous time~\cite{rosin_experiments_2013}. This approach enables a wide variety of dynamics inaccessible to synchronous digital circuits, including chaos~\cite{zhang_boolean_2009}. In ABNs, chaotic dynamics have been shown to derive from imperfections in the transistors implementing logic devices through degradation effects~\cite{cavalcante_origin_2010}. Physical implementations have been realized with individual XOR and XNOR gates with delayed feedback~\cite{rosin_experiments_2013}, a small arrangement of two XOR gates and one XNOR gate~\cite{zhang_boolean_2009}, and a ring topology of XOR gates with a single XNOR node~\cite{rosin_ultrafast_2013}. From these previous works, we establish general guidelines for constructing novel network topologies favoring sustained chaotic dynamics.

We show that a class of regular lattices of XOR and XNOR gates has properties favoring the emergence of chaos. This allows for the emergence of chaos on structures which are spatially coupled, leading to distributed Boolean chaos. We characterize the dynamics of these systems by physically realizing lattices on field-programmable gate arrays (FPGA), a digital reconfigurable chip, and performing an analysis over the observed node output waveforms. In particular, we observe fast spatial and temporal decorrelation between node signals, making the instantiated networks promising parallel entropy sources. We then explore this property for a practical application solving the MAB task.

We demonstrate here that chaotic dynamics emerging from ABNs can be used to treat MAB problems efficiently. To this end, we design a network yielding 1152 weakly dependent chaotic signals, which we use to solve MAB problems with up to 1024 arms, twice the number demonstrated in previous photonic implementations~\cite{morijiri_decision_2022}. We also show separately that convergence time scales more favorably with the number of arms than existing physical MAB accelerators. Thus, although the signal bandwidths achievable with electronic circuits are lower than the ones obtainable with optical setups, our approach allows the number of entropy channels to be effortlessly scaled, enabling massively parallel computing. Since this implementation has been fully evaluated on a commercially available reconfigurable device, it is easily reproducible with no need for specialized fabrication.

This work is organized as follows: we first establish guidelines for creating networks of clockless logic gates exhibiting chaotic dynamics. From these principles, we define a lattice structure to be studied, showing it has properties that favor the emergence of Boolean chaos. We then implement this network experimentally and characterize the signals it yields, showing they display the characteristics of Boolean chaos~\cite{zhang_boolean_2009} and demonstrating their adequacy as parallel entropy sources. Finally, we apply these signals to MAB problems and analyse the convergence time scaling as a function of the number of arms.

\section*{Results}
\subsection*{Physical Boolean lattices for complex autonomous dynamics}
To achieve a sustained chaotic regime with an autonomous Boolean network, we adopt three guidelines, inspired by the ones adopted in the construction of the architecture proposed by Rosin et al.\cite{rosin_dynamics_2015}

Firstly, we aim to construct a network only with XOR and XNOR gates; this choice is based on the abundant evidence of chaotic dynamics emerging from networks constructed with these gates\cite{zhang_boolean_2009, cavalcante_origin_2010, rosin_ultrafast_2013}, as well as isolated gates with delayed feedback\cite{ghil_boolean_1985, rosin_experiments_2013}. This choice is also reinforced by their Boolean sensitivity when analyzed in their clocked (synchronous) mode of operation\cite{kauffman_metabolic_1969, aldana_boolean_2003}. Given a Boolean function $f: \{0, 1\}^n \rightarrow \{0, 1\}$, we define the Boolean sensitivity of $f$ at a state $x \in \{0, 1\}^n$ as the number of single-bit flips (Hamming neighbors) of $x$ that yield a change in the output $s(f, x) = \sum_{i=1}^{n}\chi[f(x) \neq f(x+e_i)]$, where $\chi[A]$ is the indicator function that is equal to 1 if $A$ is true and 0 otherwise, and $e_i$ is the $i\textrm{-th}$ standard basis vector; the average sensitivity is defined over the state space $s(f) = \mathbb{E}_x[s(f,x)$]~\cite{shmulevich_activities_2004}. With this definition, it is clear that XOR and XNOR gates posses maximum average sensitivity, since any single-bit flip in any possible state yields a change in the output, which would favor the emergence of complex dynamics in a clockless operating regime.

Secondly, we take into consideration hardware constraints. Implementing gates on a FPGA establishes a constraint on the maximum gate fan-in. Considering that we aim to exploit the dynamical properties of look-up tables in these devices, a gate implementation should not be distributed among multiple look-up tables to optimize hardware resource usage; hence, we constrain the maximum fan-in to the fan-in of the FPGA's look-up tables. Moreover, we aim to implement a lattice structure with predominantly local connections, diminishing routing complexity.

Lastly, we require the network to contain no fixed points in its synchronous discrete-time Boolean representation. This criterion is relevant to the autonomous dynamics because any stationary state of the physical ABN must satisfy the same Boolean consistency conditions as a fixed point of the corresponding discrete network: if all node states become time-independent, propagation delays no longer affect their Boolean values and every node should simultaneously satisfy the Boolean function implemented by its gate. Consequently, the absence of fixed points in the discrete Boolean representation excludes stationary Boolean states of the corresponding autonomous network described by Boolean delay equations\cite{ghil_boolean_1985}. The converse is not necessarily true: the existence of a Boolean fixed point does not guarantee that the autonomous system will converge toward it, since its dynamics depend continuously on propagation delays. In a physical implementation, the dynamics are also influenced by the nonideal, finite-bandwidth response of the logic elements of the reconfigurable chip.

Previous studies of autonomous Boolean circuits provide further guidance for promoting complex (chaotic) dynamics beyond the exclusion of stationary states. In particular, XOR and XNOR gates used in our systems have been extensively used to generate Boolean chaos because of their maximum Boolean sensitivity. In autonomous operation, Boolean transitions propagate through the network according to the physical propagation delays of the logic elements and interconnections. The interplay between delayed feedback and nonideal gate dynamics, including finite bandwidth, short-pulse filtering, and history-dependent propagation delays, has been shown theoretically and experimentally to produce chaotic dynamics in XOR/XNOR-based ABNs~\cite{zhang_boolean_2009,cavalcante_origin_2010,rosin_experiments_2013}. Therefore, we combine these established properties --- high Boolean sensitivity, autonomous delayed interactions, and the absence of stationary Boolean states --- as design principles for constructing networks favorable to sustained chaotic dynamics.

\begin{figure}[t!]
\centering
\includegraphics[width=0.82\linewidth]{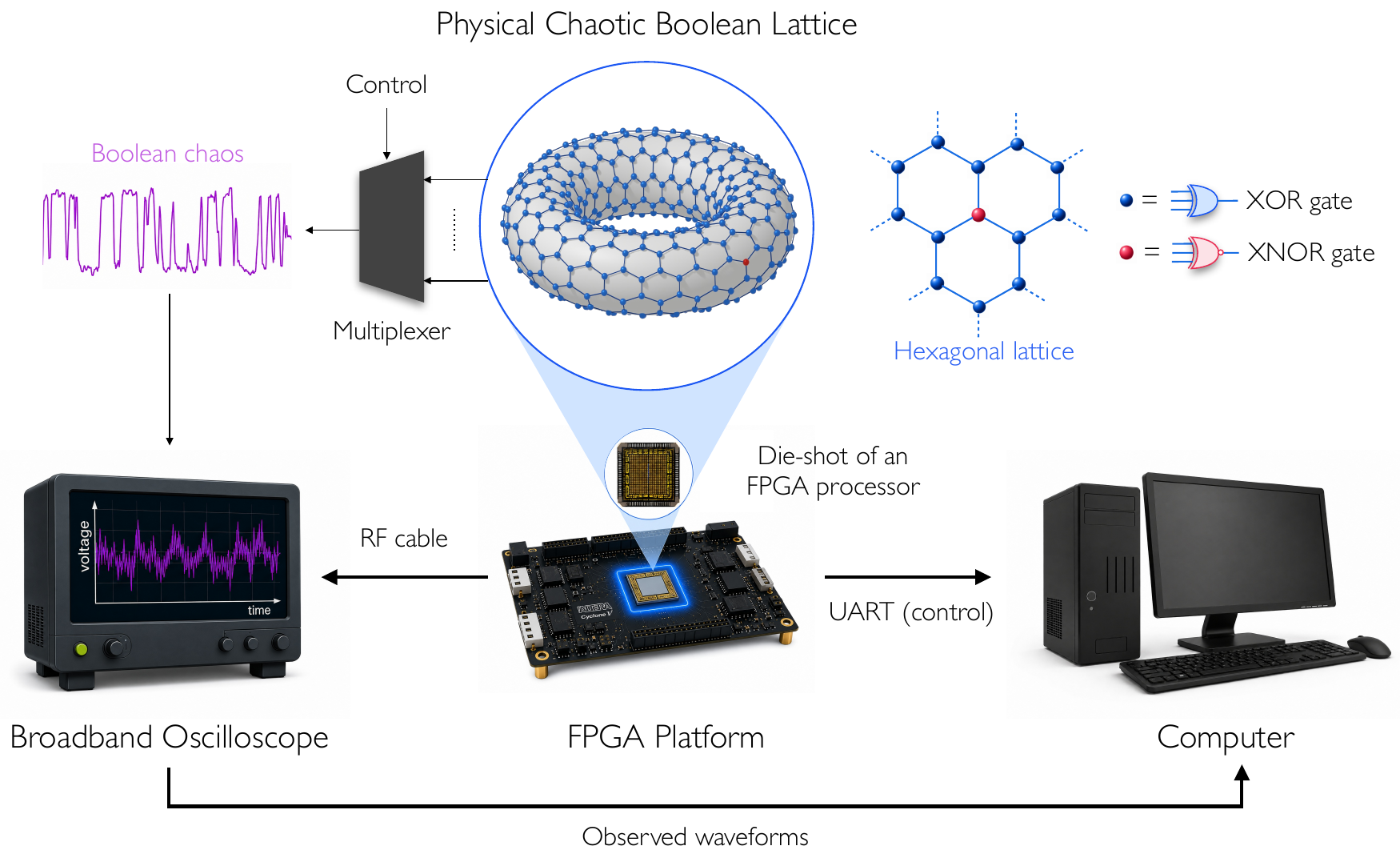}
\caption{\textbf{Boolean lattice topology and experimental setup.} The adopted topology is a periodic hexagonal lattice, constructed from bidirectionally coupled XOR gates (blue), with a single XNOR gate (red) introduced to drive instability. The lattice is represented on a torus, highlighting the periodic boundary conditions of the lattice. We also represent the repeating hexagonal pattern on the flattened graph. The lattice is implemented on an FPGA, with individual nodes measured using a high-speed oscilloscope after a 10~dB attenuation stage. Nodes to be sampled are selected via control signals from a host computer through a UART interface, and oscilloscope measurements are transferred to the host computer for offline processing.}
\label{fig:figure1}
\end{figure}

Following these design principles, we settle for the study of periodic hexagonal lattices, composed exclusively of XOR and XNOR gates. An example of one such lattice is illustrated in Fig.~\ref{fig:figure1}, with the repeating hexagonal pattern highlighted. The lattice is periodic in the sense that its boundaries are periodic, with connections established between boundary nodes in opposite sides of the graph. A convenient way to visualize this periodic connectivity structure is by projecting its graph onto a torus, which possesses an inherently periodic topology, as shown in Fig.~\ref{fig:figure1}.

The first design principle is satisfied, as we restrict the design to XOR and XNOR gates. Likewise, the second design principle is respected, since the fan-in is equal to 3 for all network nodes and a local connectivity structure is adopted for most of the lattice, with the exception of the connections at the boundaries necessary for the lattice to be periodic. Lastly, we must determine if the lattice contains fixed points, which is made viable by analysing the system as a XOR-SAT problem~\cite{moore2011nature}.
Finding a synchronous fixed point is equivalent to finding an expression such that the Boolean expressions corresponding to each of the gates constituting the network are simultaneously satisfied. Hence, solving for a fixed point is equivalent to solving a Boolean satisfiability problem (SAT). Although the general SAT problem is NP-hard; by restricting the network nodes to XOR and XNOR gates, we only need to solve a relatively easier XOR-SAT problem.

We approach this problem by modeling the network as a linear system over the finite field with two element $\mathrm{GF}(2)$. For an ABN that can be represented by a graph, the system is described by an adjacency matrix $C$ and a vector $\mathbf{b}$, whose elements $b_i$ are equal to 1 if the node at the corresponding index $i$ is an XNOR gate, and 0 if it is an XOR gate. A fixed point $\mathbf{x^*}$ is the solution to the system $(C^T + I) \mathbf{x^*} = \mathbf{b}$. Thus, we can study the existence of fixed points by analysing the properties of $A\mathbf{x^*} = \mathbf{b}$, with $A = (C^T + I)$. More details on fixed point analysis for networks composed of XOR and XNOR gates are provided in the Methods sections.

Two key properties can be inferred from this. First, if the matrix $A$ has full rank, the system has a unique fixed point, regardless of the choice of XOR and XNOR gates. Second, the network must contain at least one XNOR gate; otherwise the system has the trivial fixed point $\mathbf{x^*} = \mathbf{0}$. Together, these two properties establish a procedure for instantiating Boolean networks without a fixed point: define the network topology such that $A$ is rank-deficient, and assign XOR and XNOR gates to nodes such that the resulting system $A\mathbf{x^*} = \mathbf{b}$ is inconsistent. This inconsistency can be verified in cubic time through Gaussian elimination.

Regular periodic lattices are a natural choice for rank-deficient $A$, since they are $d$-regular networks; that is, networks with all nodes possessing both in-degree and out-degree $d$. We show that, for a $d$-regular network, the existence of a fixed point depends on the parity of $d$. Each row of $A = (C^T + I)$ sums to $d+1$. Thus, an all-ones vector satisfies $A\mathbf{1} = (d+1)\mathbf{1} = \mathbf{0}$ if $d$ is odd. If $\mathbf{1} \in \textrm{ker}(A)$, $A$ has a non-trivial null space, and is thus rank-deficient as a consequence of the fundamental theorem of linear maps~\cite{axler2024linear}. For an ABN which can be described by an undirected graph, that is, one with only bidirectional connections between nodes, $C$ is symmetric, and so is $A$ ($A = A^T$). If a solution $\mathbf{x^*}$ exists, and $d$ is odd, then $\mathbf{1}^T\mathbf{b} = \mathbf{1}^T A \mathbf{x^*} = (A \mathbf{1})^T x^* = 0$. This implies that $\sum_i b_i = 0 \pmod 2$ is a necessary condition for the existence of a fixed point. Consequently, an undirected, odd-degree lattice with an odd number of XNOR gates is proven to have no fixed points.

Any periodic hexagonal lattice has $d=3$, which suffices to ensure rank-deficiency. The same is true for any $d$-regular graph with odd $d$. We adopt lattices with only bidirectional couplings, such that they can be described by an undirected graph; hence, it suffices to instantiate the network with an odd number of XNOR gates to ensure that the system has no fixed points. Therefore, the proposed periodic hexagonal lattice respects the guidelines established for this study if it is instantiated with an appropriate number of XNOR gates.

We experimentally validated this method of analysis by implementing hexagonal and square lattices with and without synchronous fixed points; for periodic hexagonal lattices, instances with fixed points were set up by constructing them with only XOR nodes. For all lattices, we only used XOR and XNOR gates. Networks with fixed points exhibited a wide variety of behavior: all periodic hexagonal lattices displayed rapid convergence to predicted fixed points, while square lattices and periodic square lattices ranged from rapid convergence to fixed points to transient and sustained aperiodic signals. The presence of transient chaos is consistent with previously reported results for ABNs with fixed points~\cite{dhuys_super-transient_2016}. In contrast, all lattices without fixed points yielded sustained aperiodic dynamics. Consequently, the proposed design guidelines provide a strong empirical indication of whether the system will generate complex dynamics. However, the analysis developed here is founded mainly on the network's behavior in discrete time, so an analytical proof of chaos for large-scale ABNs remains an open problem. A possible path for future work is the study of large-scale chaotic ABNs through continuous-time models~\cite{glass_ordered_1998}.

Having proven that periodic hexagonal lattices of XOR and XNOR gates possess the desired properties favoring chaotic dynamics, while also having a well-defined spatial structure, we study their continuous-time dynamical behavior in an FPGA physical implementation. By convention, we implement all networks with a single XNOR gate, attributed to node number 1.

\subsection*{Characterization of distributed Boolean chaos}
Having proven that any periodic hexagonal lattice can satisfy the established guidelines, we physically implement this topology in an FPGA in order to characterize it. For reproducibility, lattices are implemented on fixed positions in the FPGA's grid, as detailed in the Supplementary Material. The experimental setup used for both dynamical system characterization and decision-making implementation is illustrated in Fig.~\ref{fig:figure1}. The network operates autonomously on FPGA, and the nodes to be measured are selected through a high-speed multiplexer included in the design. Selected signals are retrieved in an oscilloscope, and transferred to a host computer for analysis. The sampling procedure is controlled by the host computer through a UART interface. Further details on the experimental setup are provided in the Methods section.

\begin{figure}[t!]
\centering
\includegraphics[scale=1]{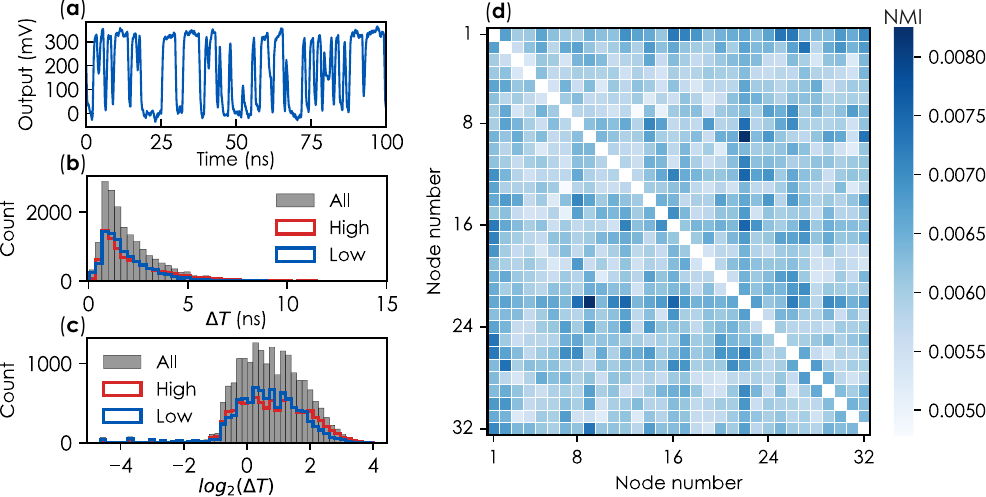}
\caption{\textbf{Properties of the measured chaotic signals.} \textbf{(a)} Measured voltage as a function of time over a 100~ns window. \textbf{(b)} Histogram of pulse lengths $\Delta T$ for logic-high (red), logic-low (blue), and combined (gray) states, measured over a 40~$\mu\text{s}$ window. \textbf{(c)} Same as (b), for $\log_2(\Delta T)$. \textbf{(d)} Pairwise normalized mutual information (NMI) computed from experimentally measured dwell-time distributions of Boolean transitions across the 32-node hexagonal lattice with periodic boundary conditions. Entropies are estimated from empirical probability distributions obtained using histogram binning with $32$ bins. }
\label{fig:figure2}
\end{figure}

We first study a network composed of 1152 nodes, arranged as a 24x24 periodic hexagonal lattice, with two nodes per grid unit. Fig.~\ref{fig:figure2}(a) illustrates a measured signal sample over a 100~ns. An initial observation suggests that the signal presents aperiodic oscillations between the logic high and logic low voltages, which is consistent with the behavior of existing chaotic ABNs~\cite{rosin_ultrafast_2013, zhang_boolean_2009}. During a 1-hour observation window, the system displayed sustained aperiodic oscillations, with no indication of collapse into periodic or fixed-point behavior. To further characterize the signal over the network, all node output signals are sampled over $40~\mathrm{\mu s}$ windows for a characterization of their statistical properties. Since acquisitions are performed over the full lattice, we report the means and population standard deviations of metrics over all nodes.

First, we calculate the power spectral densities (PSDs) for each node output signal, which is then used to infer the bandwidth until the $10~\mathrm{dB}$ dropoff point of each node. We observe that all nodes present similar flat PSDs, with bandwidth equal to $555.09 \pm 16.78~\mathrm{MHz}$. This broadband characteristic attests the aperiodicity of the signals, with bandwidth magnitude compatible with chaotic ABN literature~\cite{rosin_ultrafast_2013, zhang_boolean_2009}. The observed broadband spectra correspond, via the Wiener-Kinchin theorem, to autocorrelation functions with short correlation times, providing strong evidence of the signal's aperiodicity. This result is also compatible with previous evidence of chaotic behavior in ABNs~\cite{rosin_ultrafast_2013, zhang_boolean_2009}. Visual inspection of estimated PSDs and autocorrelation functions reveals no substantial differences among nodes, as shown in the Supplementary Material. This property motivated the estimation of means and standard deviations over all nodes, as well as further analysis over a representative node.

We further characterize the generated signals using nonlinear time-series analysis, performing noise titration~\cite{Barahona2001}, correlation dimension analysis~\cite{Grassberger1983a}, and a $0-1$ test for chaos~\cite{Toker2020} on a representative node. Noise titration gives a noise limit of $NL = 144\%$, providing evidence of nonlinear determinism in the system's dynamics. The correlation dimension analysis provides a correlation-dimension estimate of $D_c \approx 1.3 \pm 0.1$, which is consistent with a low-dimensional non-integer attractor for the reconstructed dynamics of a single measured node. Lastly, the $0-1$ test produces a value of $K=0.998$, above the adopted threshold of $0.99$, which is consistent with chaotic dynamics. Together, these results provide convergent evidence that the studied ABN lattice dynamics are chaotic (for more details, see the Supplementary Material).

From a theoretical perspective, it has been shown that the complexity observed in chaotic ABNs originates from the Boolean transition times~\cite{cavalcante_origin_2010}. Thus, since we aim to exploit ABNs as entropy sources, it is natural to study the properties of this physical quantity. We characterize the Boolean transition intervals by analysing the properties of the observed logic-high and logic-low pulses, obtained by applying a threshold at half of the maximum observed voltage. The distribution of pulse widths $\Delta T$ over $40~\mathrm{\mu s}$ sampled at $25~\mathrm{GS/s}$ yields the histogram illustrated in Fig.~\ref{fig:figure2}(b); the distributions for low and high pulses are remarkably similar, which motivates their combination into a single data stream for use as an entropy source. We further observe that the obtained distribution is positively skewed, with shorter pulses occuring more often than long pulses, which are distributed across a long tail.
Taking $\mathrm{log}_2(\Delta T)$ over the same analysis window yields the distribution modeled by the histogram in Figure \ref{fig:figure2}(c), which displays reduced skew compared to the distribution of $\Delta T$. We use this transformed signal for MAB tasks because it compresses the scale of large values in the distribution of $\Delta T$, reducing the disproportionate influence of outliers on TOW's decision-making.

To further characterize this signal, we construct the time series $T_i$ for the $i$-th node, where entries are the logarithms of pulse widths by order of occurrence (encompassing both logic-low and logic-high pulses), \emph{i.e.} $T_i=\log_2 (\Delta T_i)$. We adopt a base of 2 for the logarithm due to its ease of implementation in digital logic, since $\lfloor \log_2 (\Delta T_i) \rfloor$ corresponds directly to the position of the most significant non-zero bit of an unsigned integer representation of the pulse length. The entries of $T_i$ are generated asynchronously, continuously in time, at an average rate of $469.86 \pm 19.34~\textrm{MS/s}$. Thus, the 1152-node lattice yields an aggregate sample generation rate for $T_i$ of $0.54~\textrm{TS/s}$. To demonstrate this approach scales further, we instantiate multiple lattices on the same chip (see Supplementary Material). A grid consisting of 10 independent 512-node lattices (5120 nodes total) achieves a per-node sample generation rate for $T_i$ of $417.50 \pm 23.06~\textrm{MS/s}$, corresponding to an aggregate rate of $2.14~\textrm{TS/s}$.

To exploit the logarithmic dwell times from the various nodes of the Boolean lattice as parallel multi-channel entropy sources, it is necessary that the sequence $T_i$ satisfy two properties: (i) successive entries of $T_i$ should exhibit minimal statistical temporal dependence, and (ii) $T_i$ should exhibit weak statistical dependence with the corresponding sequences $T_j$ generated by other nodes in the network. Due to the nonlinear dynamics of the system, we quantify statistical dependencies using the normalized mutual information (NMI), defined as~\cite{kvalseth_normalized_2017}
\begin{equation}
    \mathrm{NMI}(T_i,T_j)=\frac{I(T_i,T_j)}{\min\!\left(H(T_i),H(T_j)\right)},
\end{equation}
where $I(T_i,T_j)=H(T_i) +H(T_j)-H(T_i,T_j)$ is the mutual information between the random variables $T_i$ and $T_j$, $H(T_i)$ and $H(T_j)$ are their marginal entropies, and $H(T_i,T_j)$ is the entropy of their joint distribution. The NMI values are in the range $[0,1]$ with $0$ indicating statistical independence and $1$ corresponding to complete dependence.

Temporal dependence is assessed by computing the NMI between successive logarithmic dwell times of a given node, \emph{i.e.} $\textrm{NMI}(T_i[t], T_i[t+\tau])$, for $\tau \in \left\{1, 2, \dots, 100\right\}$. Across all nodes and delays, we observe NMI levels not exceeding $0.017$, thus providing evidence of very weak statistical dependence. Similarly, we evaluate the pairwise statistical dependence between nodes for a lattice with $N$ nodes. The NMI values form a symmetric $N \times N$ matrix, with entry $\textrm{NMI}_{ij}$ quantifying the normalized statistical dependence between nodes $i$ and $j$. To maintain the computational tractability of analysis, we restrict ourselves to a $4 \times 4$ periodic hexagonal lattice made of $N=32$ Boolean nodes. We estimated this matrix over all pairs of extracted pulse-width signals. Values on the diagonals were not included in this analysis, since $\textrm{NMI}(T_i, T_i)=1$ by definition. The resulting matrix is graphically represented in Fig.~\ref{fig:figure2}(d), indicating a remarkably low mutual information ($\text{NMI} < 0.01$) for all nodes pairs. This result provides empirical evidence of the negligible pairwise statistical dependence of the Boolean-transition logarithmic dwell-time signals generated by the nodes of the ABN. This supports their intended use as parallel physical entropy sources that can be leveraged for solving MAB problems, as described in the following section.

\subsection*{Reinforcement learning with chaotic signals}
Having validated the properties of the studied ABN as an entropy source, we next investigate its application to reinforcement learning by coupling it with the TOW algorithm. Specifically, we address the multi-armed bandit (MAB) problem.

In the MAB problem, an agent seeks to maximize the reward obtained by performing a sequence of actions. The only action available to the agent is to select one of $M$ slot machines (arms) to play, each with an unknown win probability. Thus, the problem is defined by the number of machines $M$ and their win probabilities $P_i,\, i \in \{1, \ldots, M\}$, and the goal of the algorithm employed by the agent is to maximize the total reward in a limited number of plays.

\begin{figure}[t!]
\centering
\includegraphics[scale=1]{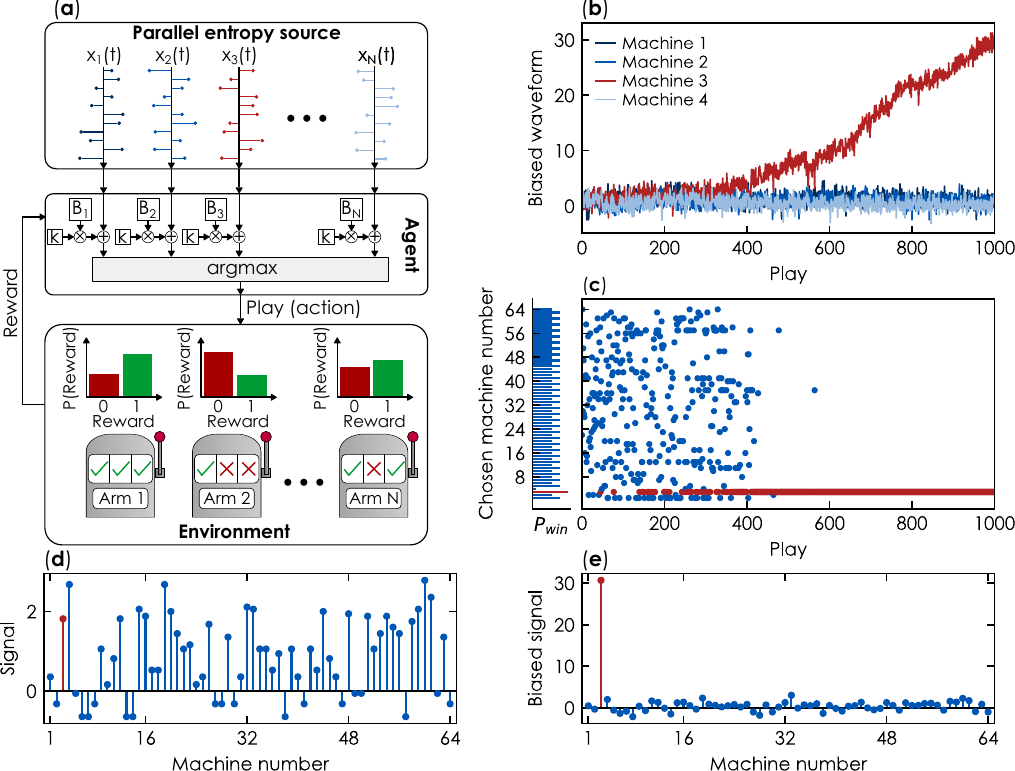}
\caption{\textbf{Solving the MAB problem using Boolean chaos.} \textbf{(a)} Decision-making scheme based on parallel chaotic entropy sources. \textbf{(b)} Biased signal produced by the TOW algorithm for a 64-armed bandit problem. The machine with the highest value is selected. \textbf{(c)} Machine selection over successive plays for one realization of the 64-armed bandit problem, alongside relative reward probabilities assigned to each machine (left), with the optimal machine highlighted in red. \textbf{(d)} Unbiased signals from the Boolean chaotic entropy source at play 1000. \textbf{(e)} Signal from (d) after bias adjustment by the TOW algorithm, at the same play.}
\label{fig:figure3}
\end{figure}

The TOW algorithm handles the exploration-exploitation trade-off~\cite{sutton2018reinforcement}, using an entropy source to drive exploration. To implement this algorithm, we employ the decision-making pipeline illustrated in Fig.~\ref{fig:figure3}(a).
The starting point of the decision process is a parallel entropy source, corresponding here to the signals $T_i,\, i \in \{1, \ldots, 1152\}$ retrieved from the experimental realization of the studied 1152-node ABN. In order to increase the number of transitions observed per oscilloscope acquisition, the acquisition window is increased from $40~\mu\textrm{s}$ to $200~\mu\textrm{s}$, and the sampling rate is reduced from $25~\textrm{GS/s}$ to $6.25~\textrm{GS/s}$; this rate remains well above the Nyquist limit for the raw Boolean chaotic signal bandwidth ($555.09~\textrm{MHz}$). Each signal is then biased by an adjustable parameter $B_i$ multiplied by a fixed scaling factor $k$, such that the biased signal is $A_i[t] = T_i[t] + k B_i[t]$. At each signal index $t$, a play is performed, with the machine having the highest biased signal $A_i[t]$ being selected. Depending on the observed outcome (win or loss), the bias parameters $B_i$ are adjusted in accordance with the TOW method. Consistent with the binary nature of outcomes, we define a reward of one for a win and zero for a loss. A detailed description of the TOW algorithm implementation is provided in the Methods section.

To compare MAB algorithms, for a problem with $M$ slot machines, we construct a benchmark with fixed win probabilities: $P_1=0.7$, $P_2=0.5$, $P_3=0.9$, $P_4=0.1$, and $P_{2j-1}=0.7$, $P_{2j}=0.5$, for $j = 3 , \ldots , \lceil M/2 \rceil$. These fixed probabilities follow prior work in the TOW literature~\cite{morijiri_parallel_2023, shen_harnessing_2023, naruse_scalable_2018, morijiri_decision_2022}, enabling direct comparison with existing approaches. In this setting, slot machine 3 has the highest win probability and is therefore the optimal choice.

We initially evaluate the decision-making pipeline over a MAB instance with 64 machines. The biased signals $A_i(t)$ for machines 1-4 are depicted in Fig.~\ref{fig:figure3}(b), while the selected machines and win probabilities are represented in Fig.~\ref{fig:figure3}(c). During the initial plays, we observe that the biases are small, and thus no machine is favored during selection. This is made evident in Fig.~\ref{fig:figure3}(b) until approximately play 400, after which the biased signal for machine 3 quickly increases until it is significantly higher than the others. This behavior is reflected in the machine choices presented in Fig.~\ref{fig:figure3}(c), which shows that multiple machines are selected until play 400, and the correct machine (machine 3) is consistently selected shortly after.

At play 1000, we represent the unbiased signals $T_i[1000]$ in Fig.~\ref{fig:figure3}(d), and the corresponding biased signals $A_i[1000]$ in Fig.~\ref{fig:figure3}(e). We observe that the bias adjustment procedure brings the signal for machine 3 significantly above the value range populated by the others, leading to the consistent selection of the optimal machine.

Hence, we assert that the TOW algorithm coupled with our studied chaotic ABN behaves correctly in this problem instance. In order to evaluate how this approach performs as the problem scales, we evaluate it over the same benchmark while varying the number of slot machines.

To evaluate the algorithm's performance across different problem sizes, we vary the number of slot machines $M$ as powers of 2 from 4 to 1024, i.e., $M \in \{4, 8, 16, 32, 64, 128, 256, 512, 1024\}$. For each $M$, we use the experimentally sampled signals $T_i$ for $i \leq M$. Each $T_i$ is read from node $i$ in the physically realized lattice. To compare performance, we adopt two metrics: correct decision rate (CDR)~\cite{naruse_decision_2014} and regret~\cite{auer_finite-time_2002}. Both metrics were evaluated over $100$ experiment realizations, each with a total duration of $18\,000$ plays.

\begin{figure}[t!]
\centering
\includegraphics[scale=1]{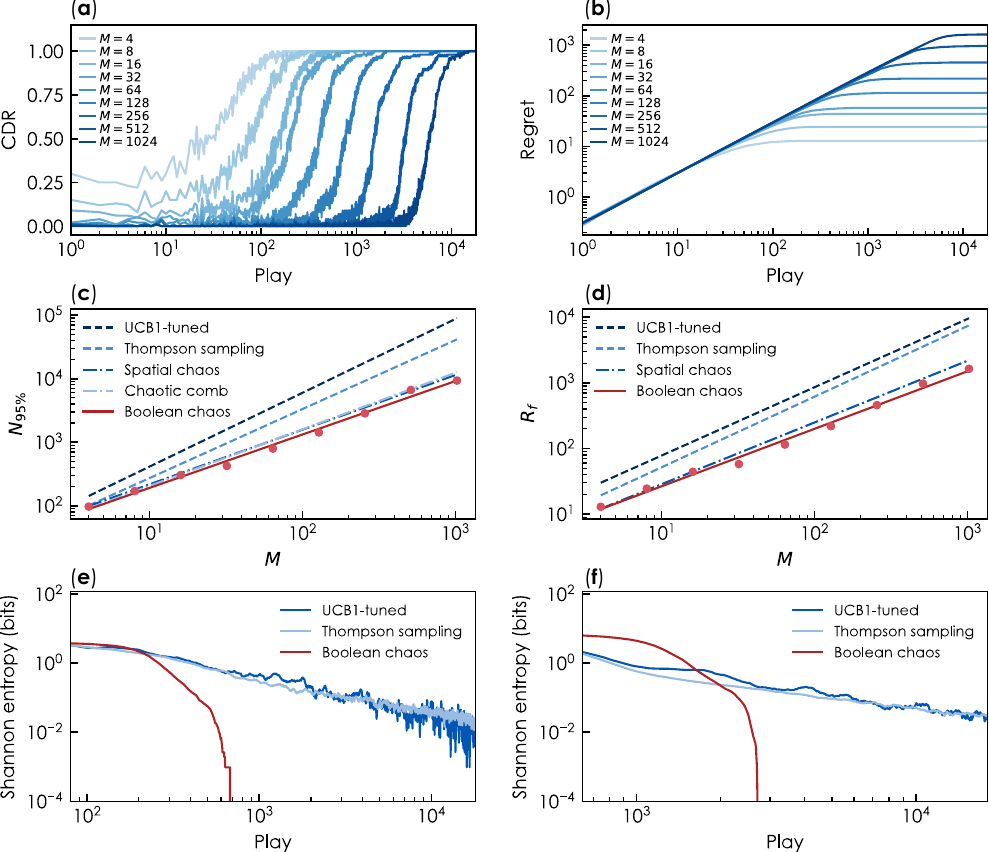}
\caption{\textbf{Scaling properties of the TOW algorithm using Boolean chaos.} \textbf{(a)} Correct decision rate (CDR) over successive plays, for problems of increasing scale. \textbf{(b)} Regret over successive plays, for problems of increasing scale. \textbf{(c)} Empirical scaling of $N_{95\%}$, the number of plays required to reach 95\% CDR, across multiple solutions to the MAB problem. \textbf{(d)} Empirical scaling of $R_f$, the regret at play $18\,000$, across different solutions to the MAB problem. \textbf{(e)} Shannon entropy of machine selection over successive plays for 16-armed bandit problem. \textbf{(f)} Same as (e), for a 128-armed bandit problem.}
\label{fig:figure4}
\end{figure}

The CDR is defined as the fraction of optimal choices taken at a given play across all experiment realizations. Figure~\ref{fig:figure4}(a) shows the CDR over successive plays for each number of slot machines $M$. For all values of $M$, a correct decision rate of one was reached; i.e., the correct machine was always chosen in last play for all $100$ experiment realizations. Furthermore, we observe that the number of iterations required for convergence to a CDR of one scales with $M$. The CDR evolution curve has a similar S-shape for all values of $M$: the CDR starts near chance level during the initial plays, then rises rapidly as evidence increasingly favors the correct decision, before saturating at its upper bound of one.

The regret metric represents a loss relative to the ideal total reward. Hence, an efficient algorithm should minimize the regret throughout the decision process. Figure~\ref{fig:figure4}(b) portrays the regret over successive plays for all problem sizes $M$. Regret initially grows linearly, then converges to a constant as the CDR saturates at one --- a direct consequence of the regret only increasing when a suboptimal decision is taken. We observe that final regret after convergence scales with the problem size $M$.

To better assess how the CDR and regret scale with the number of slot machines $M$, we assess metrics over sets of experimental realizations. Here, we evaluate $N_{95\%}$, the number of iterations for the correct decision rate to reach $0.95$, and $R_f$, the regret on the final ($18\,000$th) play.

From the obtained data points, we can infer a power law relating $N_{95\%}$ and the number of machines $M$. Figure~\ref{fig:figure4}(c) shows $N_{95\%}$ as a function of $M$ for TOW based on Boolean chaos (ours), optical spatiotemporal chaos~\cite{morijiri_parallel_2023} and microcomb-based chaos~\cite{shen_harnessing_2023}. We also show scaling laws for UCB1-tuned~\cite{auer_finite-time_2002} and Thompson sampling~\cite{thompson_likelihood_1933} applied to the MAB benchmark problem.

We infer the empirical scaling law $N_{95\%} = 27.55\,M^{0.84}$ for TOW with Boolean chaos, while the reported scaling laws for optical spatiotemporal chaos~\cite{morijiri_parallel_2023} and microcomb-based chaos~\cite{shen_harnessing_2023} are, respectively, $N_{95\%} = 30.0\,M^{0.86}$ and $N_{95\%} = 26.48\,M^{0.89}$. The scaling laws for UCB1-tuned and Thompson sampling are, respectively, $N_{95\%} = 28.49\,M^{1.16}$ and $N_{95\%} = 22.41\,M^{1.09}$, thus showing significantly worse scaling properties compared to TOW-based methods.
Therefore, our approach exhibits the smallest scaling exponent among the analysed methods, requiring the fewest iterations to converge to the correct decision as the number of slot machines grows.

Similarly, we infer a power law relating $R_f$ to the number of machines $M$. Figure~\ref{fig:figure4}(d) shows $R_f$ as a function of $M$ for TOW with Boolean chaos, along with the scaling laws for TOW with optical spatiotemporal chaos~\cite{morijiri_parallel_2023}, UCB1-tuned~\cite{auer_finite-time_2002} and Thompson sampling~\cite{thompson_likelihood_1933}.

We estimate the scaling law $R_f = 3.54\,M^{0.88}$ for TOW with Boolean chaos, while for optical spatiotemporal chaos~\cite{morijiri_parallel_2023} a scaling law $R_f = 3.29\,M^{0.94}$ was reported. UCB1-tuned and Thompson sampling present significantly worse scaling behavior, with $R_f = 7.13\,M^{1.04}$ and $R_f = 4.36\,M^{1.08}$. Hence, our approach presents the best scaling behavior among analyzed methods; that is, the final regret is the smallest as the problem size grows.

To further analyze the statistical properties of the decision-making process, we study the Shannon entropy over the choices taken at each play. For a problem with $M$ slot machines, we estimate the entropy over a moving window containing the $5M$ last plays, averaged over 100 experiment realizations. The entropy quantifies how concentrated the selection is on a single machine or a group of machines, ranging from a maximum when every slot machine is selected with equal probability to a minimum of 0 when a single machine is always selected.

Figures~\ref{fig:figure4}(e)-(f) show the Shannon entropy over successive plays for TOW with Boolean chaos, UCB1-tuned and Thompson sampling, for $M=16$ and $M=128$, respectively. In both cases, the entropy of UCB1-tuned and Thompson sampling decrease gradually as the number of plays increases, whereas the entropy of the TOW method with Boolean chaos exhibits a different profile: a slow initial decrease followed by a sharp drop, attaining low entropy substantially faster than UCB1-tuned and Thompson sampling. A similar sharp drop in entropy has been observed for a different chaos-based TOW implementation based on laser dynamics~\cite{iwami_controlling_2022}, suggesting it may be a general signature of chaos-driven exploration in TOW methods. This rapid decrease in entropy is also linked to the faster convergence to the optimal choice compared to the MAB baselines, shown in Fig.~\ref{fig:figure4}(c).

\section*{Discussion}
A key advantage of Boolean chaos for practical applications is its ease of realization across substrates. Our FPGA implementation, in particular, demonstrates an easy-to-deploy approach. Configuring this device, we obtain a lattice with chaotic behavior by exploiting the physical properties of its look-up tables (LUTs). While this substrate was chosen for its convenience and wide availability, the underlying approach is not tied to it. Indeed, the same chaotic behavior is expected to manifest from the non-idealities present in other digital logic circuits, including non-reconfigurable substrates. This suggests a promising path towards dedicated integrated circuits for parallel entropy generation.

Another property of the proposed approach is its ease of scalability. We show that regular gate lattices can be scaled effortlessly while retaining the properties favoring the emergence of Boolean chaos; alternatively, multiple lattices can also be replicated in isolation to increase the total number of nodes. In both cases, the physical implementation of an entropy channel requires only a single logic gate. In the FPGA adopted in this work, each logic gate is implemented on a single LUT, so the device could ideally support up to $114\,480$ parallel entropy sources. In practice, the FPGA implementation is restricted by the device's capability to sustain high switching rates at scale, which demands significant current to drive the logic gates.

Despite this constraint, we show that the used device supports a network with a total of 5120 nodes, each generating pulse samples at an average rate of $417.50 \pm 23.06~\textrm{MS/s}$, yielding an aggregate rate of $2.14$~TS/s. We expect further scaling to be achievable on high-end FPGAs, which typically offer improved transistor performance and better thermal dissipation. Alternatively, very-large-scale integration (VLSI) systems offer further scalability, since power-delivery networks and thermal-management infrastructure can be custom-designed for the circuit's specific switching rates. Here, the straightforward scalability of the entropy source has enabled us to solve MAB instances with up to 1024 arms (machines), which represents, to the best of our knowledge, twice the scale of the largest prior chaos-based TOW demonstration~\cite{morijiri_parallel_2023}. Furthermore, we obtain a more favorable scaling law between problem size and the number of plays required for convergence, compared to existing chaos-based TOW and conventional MAB algorithms~\cite{morijiri_parallel_2023, shen_harnessing_2023}.

This scaling result demonstrates the efficiency of the proposed entropy source for the TOW algorithm; physically integrating the decision-making module, responsible for updating biases and computing decisions, remains a natural next step toward a fully hardware-based implementation. If such a device is implemented as a digital circuit, the circuit must be capable of fetching, storing and processing samples at rates close to $500~\textrm{MS/s}$ over numerous channels, which may require substantial memory and compute resources. These limitations also constrained the scale of the presented proof-of-concept, which was restricted to 1024 arms (slot machines) by our method's data-acquisition and offline processing limits. A promising prospective solution to this issue is the use of novel analog devices for the decision making procedure; proposals for TOW decision-making modules have been made with CMOS analog circuits~\cite{analog_tow_2026} and atomic switches~\cite{lutz_ag2_2016, kim_decision_2016}. These solutions allow for a direct use of generated analog signals, enabling in-memory computing and bypassing the need for analog-to-digital conversion and memory storage.

Given sufficient development in auxiliary circuitry, the most direct application of the proposed system is the development of MAB accelerators. While MAB represents a fundamental, stateless instance of reinforcement learning, it has been shown to be directly applicable to a variety of real-world applications, including running recommendation systems~\cite{silva_multi-armed_2022}, improving online advertisement~\cite{schwartz_customer_2017}, managing risk in investment portfolios~\cite{huo_risk-aware_2017} and handling spectrum scheduling in wireless networks~\cite{li_multi-armed-bandit-based_2020}.

Overall, we have demonstrated that distributed Boolean chaotic dynamics that emerge from a carefully designed network can provide a scalable physical substrate for massively parallel reinforcement learning. By engineering clockless networks of XOR/XNOR logic elements, we obtained an on-chip collective chaotic state in which individual nodes generate high-bandwidth entropy streams with weak statistical dependence. Importantly, this parallelism is achieved through a one-to-one correspondence between physical logic elements and entropy channels, providing a direct route to scale the number of available physical entropy sources.

We experimentally realized this principle on a commercial FPGA using a 1152-node autonomous Boolean lattice and exploited its chaotic dynamics to solve multi-armed bandit problems with up to 1024 arms. Despite the lower bandwidth of electronic Boolean dynamics compared with photonic chaos, the high integration density of CMOS logic enables substantially more entropy channels within a compact architecture. Taken together with the observed favorable scaling of convergence time and regret, these results show that increasing the spatial parallelism of physical chaos can serve as an alternative to increasing the bandwidth of individual chaotic sources.

More broadly, our results establish a connection between the collective dynamics of large autonomous Boolean systems and their use as computational resources. The implementation on standard reconfigurable CMOS hardware further shows that this principle does not require specialized fabrication and could ultimately be transferred to dedicated integrated circuits in which chaotic entropy generation and decision processing are co-integrated. Distributed Boolean chaos, therefore, provides a route toward computing architectures in which large numbers of physical dynamical elements operate autonomously and in parallel, turning the intrinsic continuous-time dynamics of digital hardware into a computational resource.

\section*{Methods}

\subsection*{Experimental setup}
All experiments are realized on a Cyclone® IV EP4CE115 FPGA, set up on a Terasic DE2-115 development board; this device contains a total of $114\,480$ logic elements (LEs). Each logic element contains a D flip-flop for storage and a look-up table (LUT) with a fan-in of 4. Signals to be measured are routed to the High Speed Mezzanine Card (HSMC) interface, and transmitted through coaxial cables to the oscilloscope.

For the measurements presented in this work, we use a Tektronix MSO64B oscilloscope with a 6~GHz analog bandwidth. The instrument provides a maximum sampling rate of 50~GS/s in single-channel operation and 25~GS/s when multiple channels are acquired simultaneously. Three oscilloscope channels are utilized: one for a reference signal for triggering acquisitions, and two for node output signals. To ensure compatibility with the oscilloscope's input power range, all three signals are attenuated by 10 dB using $50\Omega$ fixed coaxial attenuators (Huber+Suhner, 6610\_SMA-50-3/199\_N).

An UART serial interface is used to control the FPGA experiment from a host computer. An Unitek USB to Serial RS-232 cable enables the connection from the computer USB port to the FPGAs RS-232 interface. Oscilloscope acquisitions are retrieved via the TekHSI (Tektronix High-Speed Interface) protocol over a local area network (LAN) connection.

\subsection*{Data acquisition}
We use two high-speed multiplexers, included in the circuit design, to select the node signals to be sent to the oscilloscope. The multiplexer components are placed at fixed positions in the FPGA grid, in order to minimize analog signal distortion and improve reproducibility. To mitigate the effect of pulse-width distortion caused by rise and fall time mismatch in the FPGA's look-up tables, each stage of the multiplexer inverts the selected output. If the number of stages is odd, an additional inverter is added to the last stage.

The nodes to be sampled are controlled by Python scripts, responsible for transmitting the node indices to be sampled from the host computer to the FPGA through the UART serial interface. When a selection message is sent to the FPGA, the reception is verified with a handshake message containing the nodes being measured, sent from the FPGA to the host computer. We adopt a baud rate of $115\,200$ for the UART interface.

\subsection*{Boolean fixed points}
To evaluate the existence of fixed points in a lattice composed only of XOR and XNOR gates, we seek a point satisfying the Boolean expression represented by the network. Finding such a fixed point is equivalent to solving an instance of the XOR-SAT problem~\cite{moore2011nature}. Unlike the general SAT problem, XOR-SAT is known to be solvable in cubic time ($\mathcal{O}(n^3)$, where $n$ is the number of Boolean variables); since instances of this problem class can be mapped onto linear algebra problems over a finite field, they can be solved via Gaussian elimination.

$\mathrm{GF}(2)$ is the finite field with two elements, $\{0, 1\}$, where the addition and multiplication are defined modulo 2, such that the addition coincides with the logic XOR operation and multiplication coincides with the logic AND operation. In this field, the XNOR logic operation can be modeled as an addition with a bias of 1, since $\lnot (x \oplus y) = x \oplus y \oplus 1$. Therefore, for a network with $N$ nodes and states $x_i, i \in \{1, \cdots, N\}$, let the gate types be defined by $b_i \in \mathrm{GF}(2)$, where $b_i = 0$ denotes an XOR gate and $b_i = 1$ denotes an XNOR gate. Thus the gate's output states are given by
\begin{equation}
    x_j = b_j + \sum_{i=1}^{N} C_{ij}x_i \,, \qquad j = 1, \dots, N
\end{equation}
where $C_{ij} \in \mathrm{GF}(2)$ is 1 if node $i$ is an input of node $j$, and 0 otherwise. Therefore, by grouping the state values $x_i$ and bias values $b_i$ into vectors $\mathbf{x}$ and $\mathbf{b}$, we can rewrite this problem in matrix form:
\begin{equation}
    C^T \mathbf{x} = \mathbf{x} + \mathbf{b} \iff (C^T + I) \mathbf{x} = \mathbf{b}
\end{equation}
where $C$ is the adjacency matrix of the network under study. This equation can be expressed as the matrix equation $A\mathbf{x} = \mathbf{b}$ over $\mathrm{GF}(2)$, with $A = C^T + I$, such that the system can be analysed using tools from linear algebra.

\subsection*{Tug-of-war algorithm}
Here, we adopt the formulation described by Morijiri et al.~\cite{morijiri_parallel_2023}. We define the algorithm for a MAB instance with $M$ slot machines; each machine can either produce a win or a loss, corresponding to a reward that is either a one or a zero. In accordance with the problem size, we define $M$ parallel signals read from an entropy source $X_i[t], i\in\{1, \cdots, M\}$. In this context, the signals $X_i[t]$ are the transformed pulse width time series extracted from the studied chaotic system. These signals are biased as defined in
\begin{equation}
    A_i[t] = X_i[t] + k B_i[t]
\end{equation}
where $B_i[t]$ denotes the bias given to the signal $X_i[t]$ at play $t$, and $k$ is a fixed bias coefficient. A comparison is then taken between the $A_i[t]$ signals, with the selected machine $S[t]$ being chosen based on the highest biased signal at play $t$, such that
\begin{equation}
    S[t] = \underset{i\in\{1, \cdots, M\}}{\textrm{argmax}} A_i[t]
\end{equation}

The learning procedure is established by the adjustment of the bias term: the bias for a machine is increased if a win is obtained when selecting it; otherwise, it is decreased. The bias adjustment procedure is expressed as~\cite{morijiri_parallel_2023}
\begin{equation}
    B_i[t] = Q_i[t] - \frac{1}{M-1}\sum_{i' \neq i}^M Q_{i'}[t]
\end{equation}

\begin{equation}
Q_i[t] = \Delta W_i[t] - \omega L_i[t], \; \text{where} \; \omega = \hat{P}_{\mathrm{top1}} - \hat{P}_{\mathrm{top2}}, \; \Delta = 2 - \omega
\end{equation}

\begin{equation}
    \hat{P}_i = \frac{W_i[t]}{(W_i[t] + L_i[t])}
\end{equation}
where $t$ is the play index and $i$ is the machine index; $Q_i[t]$ is the evaluation value of machine $i \in \{1, \cdots, M\}$; $W_i[t]$ and $L_i[t]$ are the total numbers of wins and losses, respectively, up to play $t$. $\hat{P}_{\textrm{top1}}$ and $\hat{P}_{\textrm{top2}}$ are, respectively, the highest and second highest estimated win probabilities in $\{\hat{P}_i, i \in \{1, \cdots, M\}\}$.

The only hyperparameter in this algorithm is the bias scaling factor $k$. We determined its optimal values via parameter sweeps, selecting the value that yields the smallest final regret for each number of machines $M$; scaling factors for which the regret does not converge are excluded. For $M \in \{4, 8, 16, 32, 64, 128, 256, 512, 1024\}$, the search was executed over the interval $[0.1, 0.7]$ for $M \leq 32$ and $[0.5, 0.9]$ for $M \geq 64$, in both cases with a step size of $0.01$. The resulting optimal scaling factors were $0.15$, $0.22$, $0.31$, $0.64$, $0.64$, $0.72$, $0.77$, $0.67$ and $0.83$ respectively.

\subsection*{Conventional multi-armed bandit algorithms}
For evaluation of the decision-making algorithm proposed here, two algorithms widely adopted in MAB literature are used as baselines: UCB1-tuned and Thompson sampling.

UCB1-tuned is an algorithm without hyperparameters, which is based on the selection of the maximum upper confidence bound metric $\textrm{UCB}_i[t]$ calculated for each machine (arm) $i$ at play $t$~\cite{auer_finite-time_2002}, as determined by
\begin{equation}
    \textrm{UCB}_i[t] = \frac{W_i[t]}{W_i[t] + L_i[t]} + \sqrt{\frac{\ln t}{W_i[t] + L_i[t]} \min \left( \frac{1}{4}, \sigma_i^2[t] + \sqrt{\frac{2 \ln t}{W_i[t] + L_i[t]}} \right)}
\end{equation}
where $W_i[t]$ is the number of wins and $L_i[t]$ is the number of losses for machine $i$ until play $t$; $\sigma_i[t]$ is the sample variance of the reward in this interval for machine $i$.

In contrast, Thompson sampling is a Bayesian approach in which a prior distribution over machine reward probabilities is updated after each play~\cite{thompson_likelihood_1933}. At each play, a sample $\theta_i[t]$ is drawn from the posterior distribution over each machine's reward probability, and the machine with the highest sampled value is selected. To account for the binary reward structure, the priors and posteriors are modeled by Beta distributions, leading to the selection process
\begin{equation}
    \theta_i[t] \sim \mathrm{Beta}\left(\alpha_i[t], \beta_i[t]\right), \qquad S[t] = \underset{i\in\{1, \cdots, M\}}{\textrm{argmax}} \theta_i[t]
\end{equation}

for $M$ machines, where $\alpha_i[t] = 1 + W_i[t]$ and $\beta_i(t) = 1 + L_i[t]$ are the shape parameters of the Beta posterior for machine $i$, and $S[t]$ is the selected machine at play $t$. We adopt an uninformative $\mathrm{Beta}(1,1)$ prior at the first play.

\subsection*{Multi-armed bandit performance metrics}
To evaluate the decision-making performance, we define metrics calculated over multiple experiment repetitions, enabling a statistical assessment of algorithm behavior. Following established conventions in TOW literature, we adopt the correct decision rate (CDR)~\cite{naruse_decision_2014} and regret~\cite{auer_finite-time_2002} as our primary comparison metrics for evaluation performance on the MAB problem.

For an experiment repeated $N$ times, the correct decision rate is defined as~\cite{morijiri_parallel_2023}
\begin{equation}
\textrm{CDR}[t] = \dfrac{1}{N}\sum_{i=1}^N C(i, t)
\end{equation}
where $C(i, t)$ is an indicator function equal to $1$ if the correct machine is selected on the $t$-th play of the $i$-th repetition, and $0$ otherwise.

In addition to this metric, we evaluate the regret metric~\cite{auer_finite-time_2002}, which indicates the accumulated difference between the ideal total reward and the one obtained with the evaluated algorithm. For an experiment repeated $N$ times, we define the regret as~\cite{morijiri_parallel_2023}
\begin{equation}
    \textrm{Regret}[t] = t\,P_{max} - \frac{1}{N}\sum_{l=1}^N\sum_{m=1}^M P_m\,S_{l,m}[t]
\end{equation}
for a problem with $M$ slot machines, where $P_m$ the reward probability of machine m, $P_{\max}$ is the maximum reward probability among the machines, $t$ is the current play number, and $S_{l,m}[t]$ is the accumulated number of selections of machine $m$ on the $l\textrm{-th}$ repetition until the $t\textrm{-th}$ play.

\section*{Data availability}
All data needed to evaluate the conclusions in the paper are present in the paper and/or the Supplementary Material.
Additional data are available from the corresponding authors upon reasonable request.

\section*{Code availability}
The code supporting the findings of this study is available from the corresponding authors upon reasonable request.

\bibliography{Biblio}

\section*{Acknowledgements}

The authors would like to gratefully acknowledge the financial support of the Conseil of Region Grand-Est, the Air Force Office of Scientific Research (AFOSR) and Office of Naval Research (ONR) through grant FA8655-22-1-7031, the Agence Nationale de la Recherche (ANR) through grant ANR-24-CE24-4767 (AATLAS), and the Intel FPGA Academic Program. 

\section*{Author contributions statement}
D.R. designed the study, conceived the experiment and managed the overall project. E.O.G was supervised by D.R for this work. E.O.G performed the experiments with the FPGA platform, processed the data, and developed the theoretical framework for the system analysis. All the authors discussed the results and analysed the data. D.R. and E.O.G. wrote the core text of the manuscript and E.O.G the methods section. All the authors reviewed and contributed to the improvement of the manuscript.

\section*{Competing interests}
The authors declare no conflicts of interest.

\end{document}